\documentclass[11pt,a4paper]{article}
\usepackage[hyperref]{eacl2021}
\usepackage{times}
\usepackage{latexsym}
\usepackage{graphicx} 
\usepackage{subcaption} 
\usepackage{comment} 

\usepackage{microtype}

\aclfinalcopy 

\title{On the Evolution of Syntactic Information Encoded by BERT's Contextualized Representations}

\author{
  Laura Pérez-Mayos\textsuperscript{1}, Roberto Carlini\textsuperscript{1}, Miguel Ballesteros\textsuperscript{2}, Leo Wanner\textsuperscript{3,1} \\[5pt]
  \textsuperscript{1} TALN Research Group, Pompeu Fabra University, Barcelona, Spain \\
  \textsuperscript{2} Amazon AI \\
  \textsuperscript{3} Catalan Institute for Research and Advanced Studies (ICREA), Barcelona, Spain \\[5pt]
  \texttt{\{laura.perezm$\mid$roberto.carlini$\mid$leo.wanner\}@upf.edu} \\
  \texttt{ballemig@amazon.com}
}

\date{}

\begin{document}
\maketitle

\begin{abstract}

The adaptation of pretrained language models to solve supervised tasks has become a baseline in NLP, and many recent works have focused on studying how linguistic information is encoded in the pretrained sentence representations. Among other information, it has been shown that entire syntax trees are implicitly embedded in the geometry of such models. As these models are often fine-tuned, it becomes increasingly important to understand how the encoded knowledge evolves along the fine-tuning. In this paper, we analyze the evolution of the embedded syntax trees along the fine-tuning process of BERT for six different tasks, covering all levels of the linguistic structure. Experimental results show that the encoded syntactic information is forgotten (PoS tagging), reinforced (dependency and constituency parsing) or preserved (semantics-related tasks) in different ways along the fine-tuning process depending on the task.

\end{abstract}

\section{Introduction}
\label{sec:intro}
Adapting unsupervised pretrained language models (LMs) to solve supervised tasks has become a widely spread practice in NLP, with models such as ELMo \citep{peters2018deep} and, most notably, BERT \citep{devlin2018bert}, achieving state-of-the-art results in many well-known Natural Language Understanding benchmarks like GLUE \citep{wang2018glue} and SQuAD \cite{rajpurkar2018know}.
Several studies investigate what the LMs learn, how and where the learned knowledge is represented and what the best methods to improve it are; cf., e.g., \cite{rogers2020primer}. There is evidence that, among other information (such as, e.g., PoS, syntactic chunks and roles \citep{tenney2018what, Lin_2019, belinkov2017neural},  morphology in general \citep{peters2018deep}, or sentence length \citep{adi2016fine}) BERT deep models’ vector geometry implicitly embeds entire syntax trees  \citep{hewitt2019structural}. However, rather little is understood about how these representations change when fine-tuned to solve downstream tasks \citep{peters2019tune}.

In this work, we aim to understand how syntax trees implicitly embedded in the geometry of deep models evolve along the fine-tuning process of BERT on different supervised tasks, and shed some light on the importance of the syntactic information for those tasks. Intuitively, we expect morpho-syntactic tasks to clearly reinforce the encoded syntactic information, while tasks that are not explicitly syntactic in nature should maintain it in case they benefit from syntax \citep{kuncoro2020syntactic} and lose it if they do not. In order to cover the three main levels of the linguistic description (morphology, syntax and semantics), we select six different tasks: PoS tagging, constituency parsing, syntactic dependency parsing, semantic role labeling (SRL), question answering (QA) and paraphrase identification. The first three inherently deal with (morpho-)syntactic information while the latter three, which traditionally draw upon the output of syntactic parsing \cite[inter-alia]{carreras-marquez-2005-introduction,bjorkelund2010high,strubell2018linguisticallyinformed,wang-etal-2019-best}, deal with higher level, semantic information. Almost all of our experiments are on English corpora; one is on multilingual dependency parsing.

\section{Related work}
\label{sec:related_work}

BERT has become the default baseline in NLP, and consequently, numerous studies analyze its linguistic capabilities in general \citep{rogers2020primer, henderson2020unstoppable}, and its syntactic capabilities in particular \citep{linzen2020syntactic}. Even if syntactic information is distributed across all layers \citep{durrani-etal-2020-analyzing}, BERT captures most phrase-level information in the lower layers, followed by surface features, syntactic features and semantic features in the intermediate and top layers \citep{jawahar-etal-2019-bert, tenney-etal-2019-bert, hewitt2019structural}. The syntactic structure captured by BERT adheres to that of the Universal Dependencies \citep{kulmizev2020neural}; different syntactic and semantic relations are captured by self-attention patterns \citep{kovaleva2019revealing, limisiewicz2020universal, ravishankar-etal-2021-attention}, and it has been shown that full dependency trees can be decoded from single attention heads \citep{ravishankar-etal-2021-attention}. BERT performs remarkably well on subject-verb agreement \citep{goldberg2019assessing}, and is able to do full parsing relying only on pretraining architectures and no decoding \citep{vilares2020parsing}, surpassing existing sequence labeling parsers on the Penn Treebank dataset \citep{de2006generating} and on the end-to-end Universal Dependencies Corpus for English \citep{silveira2014gold}. It can generally also distinguish good from bad completions and robustly retrieves noun hypernyms, but shows insensitivity to the contextual impacts of negation \citep{ettinger2020bert}.

Different supervised probing models have been used to test for the presence of a wide range of linguistic phenomena in the BERT model \citep{conneau-etal-2018-cram, liu-etal-2019-linguistic, tenney2018what, voita2020information, elazar2020bert}.
\citet{hewitt2019structural}'s structural probe shows that entire syntax trees are embedded implicitly in BERT's vector geometry. Extending their work, \citet{chi2020finding} show that multilingual BERT recovers syntactic tree distances in languages other than English and learns representations of syntactic dependency labels.

Regarding how fine-tuning affects the representations of BERT, \citet{gauthier2019linking} found a significant divergence between the final representations of models fine-tuned on different tasks when using the structural probe of \citet{hewitt2019structural}, while \citet{merchant2020happens} concluded that fine-tuning is conservative and does not lead to catastrophic forgetting of linguistic phenomena -- which our experiments do not confirm. However, we find that the  encoded  syntactic  information  is  forgotten, reinforced or preserved differently along the fine-tuning process depending on the task. 

\section{Experimental setup}
\label{sec:experimental_setup}

We study the evolution of the syntactic structures discovered during pretraining along the fine-tuning of BERT-base (cased) \citep{devlin2018bert}\footnote{Our experiments are implemented in PyTorch, using two open-source libraries: the Transformers library \citep{Wolf2019HuggingFacesTS} and AllenNLP \citep{Gardner2017AllenNLP}. Implementation details, pretrained weights and full hyperparameter values can be found in the libraries documentation.}  on six different tasks, drawing upon the structural probe of \citet{hewitt2019structural}.\footnote{We use the same experimental setup used by the authors. Source: \url{https://github.com/john-hewitt/structural-probes}} We fine-tune the whole model on each task outlined below for 3 epochs, with a learning rate of $5e^{-5}$, saving 10 evenly-spaced checkpoints per epoch. The output of the last layer is used as input representation for the classification components of each task. To mitigate the variance in performance induced by weight initialization and training data order \citep{dodge2020fine}, we repeat this process 5 times per task with different random seeds and average results.

\vspace*{0.1cm}
\noindent{\bf PoS tagging.}
We fine-tune BERT with a linear layer on top of the hidden-states output for token classification.\footnote{Source: \url{https://github.com/Tarpelite/UniNLP/blob/master/examples/run_pos.py}} Dataset: Universal Dependencies Corpus for English (UD 2.5 EN EWT \citet{silveira2014gold}).

\vspace*{0.1cm}
\noindent{\bf Constituency parsing.} Following \citet{vilares2020parsing}, we cast constituency parsing as a sequence labeling problem, and use a single feed-forward layer on top of BERT to directly map word vectors to labels that encode a linearized tree. Dataset: Penn Treebank \citep{marcus1993building}.

\vspace*{0.1cm}
\noindent{\bf Dependency parsing.}
We fine-tune a Deep Biaffine neural dependency parser \citep{dozat2016deep} on three different datasets: i) UD 2.5 English EWT \cite{silveira2014gold}; ii) a multilingual benchmark generated by concatenating the UD 2.5 standard data splits for German, English, Spanish, French, Italian, Portuguese, and Swedish  \cite{11234/1-3105}, with gold PoS tags; iii) PTB SD 3.3.0 \citep{de2006generating}.

\vspace*{0.1cm}
\noindent{\bf Semantic role labeling.} Following \citet{shi2019simple}, we decompose the task into i) predicate sense disambiguation and argument identification, and ii) classification. Both subtasks are casted as sequence labeling, feeding the contextual representations into a one-hidden-layer MLP for the first, and a one-layer BiLSTM followed by a one-hidden-layer MLP for the latter. Dataset: OntoNotes corpus \citep{weischedel2013ontonotes}.

\vspace*{0.1cm}
\noindent{\bf Question answering.}
We fine-tune BERT with a linear layer on top of the hidden-states output to compute span start logits and span end logits.\footnote{Source: \url{https://github.com/huggingface/transformers/tree/master/examples/question-answering}.} Dataset: Stanford Question Answering Dataset (SQuAD \citep{rajpurkar2018know}).

\vspace*{0.1cm}
\noindent{\bf Paraphrase identification.}
We fine-tune BERT with a linear layer on top of the pooled sentence representation.\footnote{Source: \url{https://github.com/huggingface/transformers/blob/master/examples/text-classification/run_glue.py}.} Dataset: Microsoft Research Paraphrase Corpus (MRPC) \citep{dolan2005automatically}.

\section{Evolution of syntax trees}
\label{sec:experiments}

\citet{hewitt2019structural}'s structural probe evaluates how well syntax trees are embedded in a linear transformation of the network representation space, performing two different evaluations: i) Tree distance evaluation, in which squared L2 distance encodes the distance between words in the parse tree, and ii) Tree depth evaluation, in which squared L2 norm encodes the depth of the parse tree.

Using their probe, \citeauthor{hewitt2019structural} show that the 7th layer of BERT-base is the layer that encodes more syntactic information. Therefore, to analyze the evolution of the encoded syntax trees, we train the probes on the 7th layer of the different checkpoint models generated along the fine-tuning process of each task.\footnote{Cf. also Supplementary Material.}

\subsection{Tree distance evaluation}
\label{subsec:tree_distance}

The probe evaluates how well the predicted distances between all pairs of words in a model reconstruct gold parse trees by computing the Undirected Unlabeled Attachment Score (\textit{UUAS}). It also computes the Spearman correlation between true and predicted distances for each word in each sentence, averaging across all sentences with lengths between 5 and 50 (henceforth referred to as \textit{DSpr.}).

\begin{figure}[t]
    \includegraphics[width=0.49\textwidth]{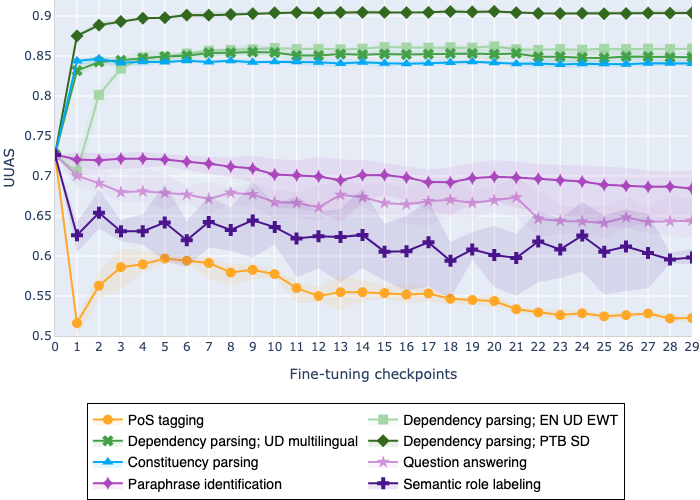}
    \caption{Tree distance evaluation. {\it UUAS } evolution.}
    \label{fig:distance_uuas}
\end{figure}

\begin{figure}[t]
    \includegraphics[width=0.49\textwidth]{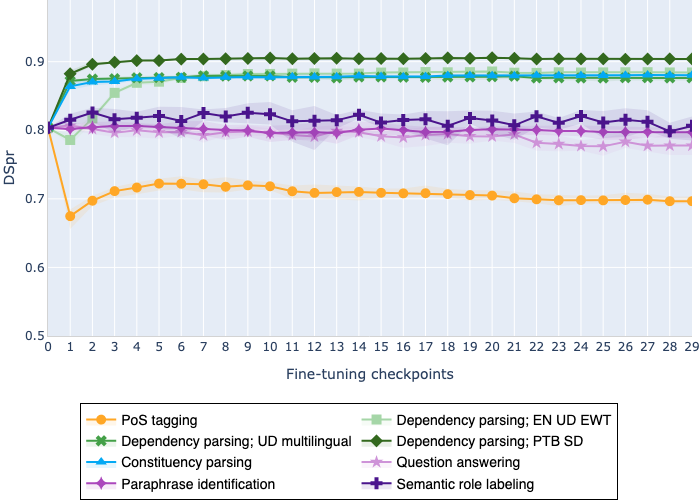}
    \caption{Tree distance evaluation. {\it Dspr } evolution.}
    \label{fig:distance_dspr}
\end{figure}

\vspace*{0.1cm}
\noindent{\bf Morpho-syntactic tasks} As shown in Figures \ref{fig:distance_uuas} and  \ref{fig:distance_dspr}, both metrics follow a similar behaviour (shades represent the variability across the 5 model runs). PoS tagging shows an important loss of performance all along the fine-tuning process, especially noticeable for UUAS (Figure \ref{fig:distance_uuas}), suggesting that distance-related syntactic information is of less relevance to PoS tagging than could be intuitively assumed.  As many words have a clear preference towards a specific PoS, especially in English, and most of the ambiguous cases can be resolved using information in the close vicinity (e.g.,  a simple 3-gram sequence tagger is able to achieve a very high accuracy \citep{manning2011part}), syntactic structure information may not be necessary and, therefore, the model does not preserve it. This observation is aligned with \citet{pimentel2020information}, who found that PoS-tagging is not an ideal task for contemplating the syntax contained
in contextual word embeddings. The loss is less pronounced on depth-related metrics, maybe because the root of the sentence usually corresponds to the verb, which may also help in identifying the PoS of surrounding words.

Constituency parsing and dependency parsing share a very similar tendency, with a big improvement in the first fine-tuning steps preserved along the rest of the process. As both tasks heavily rely on syntactic information, this improvement intuitively makes sense. Dependency parsing fine-tuned on the Penn Treebank (PTB) shows even higher results since the probing is trained on the same dataset. Interestingly, the probe performs similarly even if the parsing task is modeled as a sequence labeling problem (as in constituency parsing), suggesting that the structure of syntax trees emerges in such models even when no tree is explicitly involved in the task. The initial drop observed for PoS tagging and monolingual dependency parsing with UD, trained on UD EN EWT, may be related to the size of the dataset, since UD EN EWT is significantly smaller than the other datasets and therefore the models see less examples per checkpoint.

\vspace*{0.1cm}
\noindent {\bf Semantics-related tasks} As shown in Figures \ref{fig:distance_uuas} and  \ref{fig:distance_dspr}, both  metrics follow different behaviours (again, shades represent the variability across the 5 model runs). Paraphrase identification shows a small but constant UUAS loss along the fine-tuning, while QA shows a slightly steeper loss trend. Initially, SRL loses around 12 points, suggesting that it discards some syntactic information right at the beginning, and follows a similar downward trend afterwards. Those three tasks show a stable performance along the fine-tuning for the DSpr metric, which implies that even if there is a loss in UUAS information it does not impact the distance ordering.

\subsection{Tree depth evaluation}
\label{subsec:tree_depth}

\begin{figure}[t]
    \includegraphics[width=0.49\textwidth]{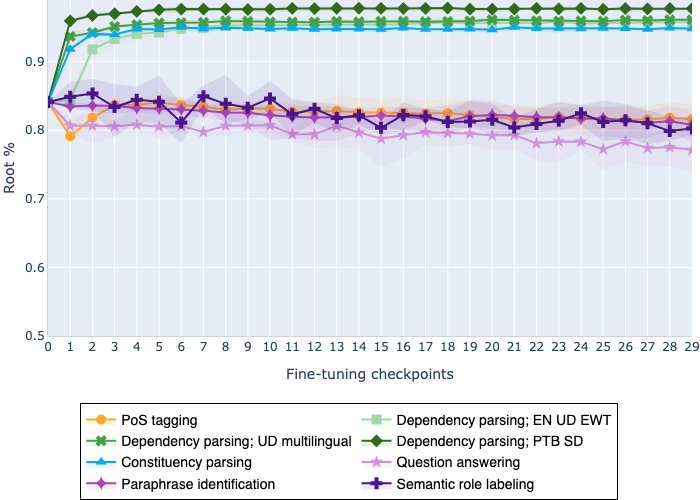}
    \caption{Tree depth evaluation. {\it Root \%} evolution.}
    \label{fig:depth_root}
\end{figure}

The probe evaluates models with respect to their ability to recreate the order of words specified by their depth in the parse tree, assessing their ability to identify the root of the sentence as the least deep word (\textit{Root \%}) and computing the Spearman correlation between the predicted and the true depth ordering, averaging across all sentences with lengths between 5 and 50 (henceforth referred  to as \textit{NSpr}).

\begin{figure}[t]
    \includegraphics[width=0.49\textwidth]{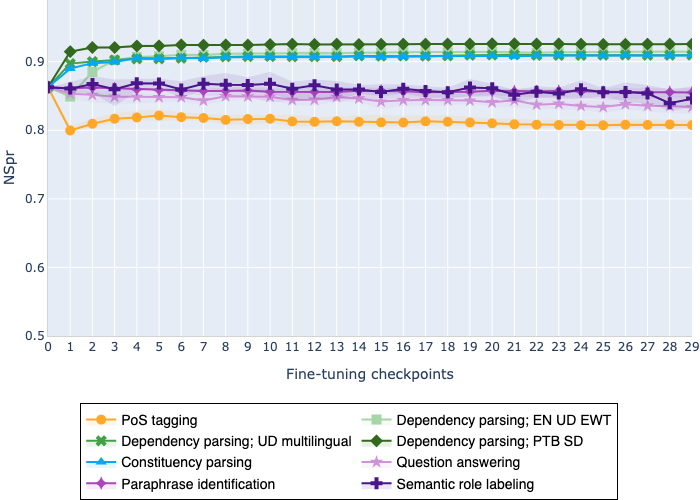}
    \caption{Tree depth evaluation. {\it Nspr } evolution.}
    \label{fig:depth_nspr}
\end{figure}

\vspace*{0.1cm}
\noindent{\bf Morpho-syntactic tasks} Again, both metrics follow a similar behaviour, as shown in Figures \ref{fig:depth_root} and \ref{fig:depth_nspr}. PoS tagging shows a sustained loss of performance, though softer than the loss observed for the distance metrics. This loss is slightly less pronounced for {\it Root \%} than for {\it Nspr}, suggesting that while depth-related syntactic information may be of less relevance to PoS tagging than it is to the other morpho-syntactic tasks, identifying the root of the sentence may be important, as the root of the sentence is likely to become one of the ambiguous tags and therefore identifying it may help to select the correct label. Constituency parsing and dependency parsing share a similar tendency, with a big improvement in the first steps preserved along the rest of the fine-tuning process, reinforcing the intuition previously introduced in Section \ref{subsec:tree_distance} about the structure of syntax trees emerging in models even when no tree is explicitly involved in the task. Again, an initial drop can be observed for PoS tagging and monolingual dependency parsing with UD, most probably related to the smaller size of the UD EN EWT dataset used in both tasks.

\vspace*{0.1cm}
\noindent{\bf Semantics-related tasks} Both metrics follow a similar behaviour, as shown in Figures \ref{fig:depth_root} and \ref{fig:depth_nspr}, with all tasks following a soft but sustained loss of performance until the end of the fine-tuning process, specially noticeable for \textit{Root \%}.

\section{Conclusions}
\label{sec:conclusions}

We show that fine-tuning is not always a conservative process. Rather, the syntactic information initially encoded in the models is forgotten (PoS tagging), reinforced (parsing) or preserved (semantics-related tasks) in different (sometimes unexpected) ways along the fine-tuning, depending on the task. Thus, we expected that morpho-syntactic tasks  clearly reinforce syntactic information. However, PoS tagging forgets it, which, on the other side, can also be justified linguistically (cf. Section \ref{subsec:tree_distance}).
In contrast, tasks closer to semantics mostly preserve the syntactic knowledge initially encoded. This interesting observation reinforces recent findings that models benefit from explicitly injecting syntactic information for such tasks \citep{singh2020syntax}.

Overall, we observed that morpho-syntactic tasks experiment substantial changes in the initial phases, while semantic-related tasks maintain a more stable trend, highlighting the importance of syntactic information in tasks that are not explicitly syntactic in nature \citep{kuncoro2020syntactic}. These observations lead to some interesting insights, but also to further questions; for instance: Can we find a specific set of probes covering different linguistic phenomena to be used as a pretraining stopping criteria? Would this lead to an improvement in the encoding of the linguistic information on pretrained models?

\section*{Acknowledgments}
This work has been partially funded by the European Commission via its H2020 Research Program under the contract numbers 779962, 786731, 825079, and 870930.

\bibliography{eacl2021}
\bibliographystyle{acl_natbib}

\appendix

\section{Target tasks performance evolution}
\label{sec:supplemental}

To complement the results shown in the main paper, we include here the performance curves of the target tasks for which the models are fine-tuned, along with the performance curves of the structural probes metrics, facilitating the comparison of the evolution of the encoded syntax trees information and the target tasks performances.

\begin{figure*}
    \centering
    \begin{subfigure}[t]{0.49\textwidth}
        \vskip 0pt
        \paragraph{PoS tagging} reaches a 0.95 accuracy in only two checkpoints, ending up with a 0.97 on the last checkpoint (Figure \ref{fig:PoS_Tagging_pos_acc}). It shows a loss of accuracy for the four probing metrics all along the fine-tuning process, especially noticeable for \textit{UUAS} (Figure \ref{fig:PoS_Tagging_distance_uuas}) and \textit{Root \%} (Figure \ref{fig:PoS_Tagging_depth_root}), suggesting that syntactic information is of less relevance to PoS tagging than could be intuitively assumed. The loss is less pronounced on depth-related metrics, maybe due to the fact that the root of the sentence usually corresponds to the verb, which may also help in identifying the PoS of surrounding words.
    \end{subfigure}
    ~
    \begin{subfigure}[t]{0.49\textwidth}
        \vskip 0pt
        \includegraphics[width=\textwidth]{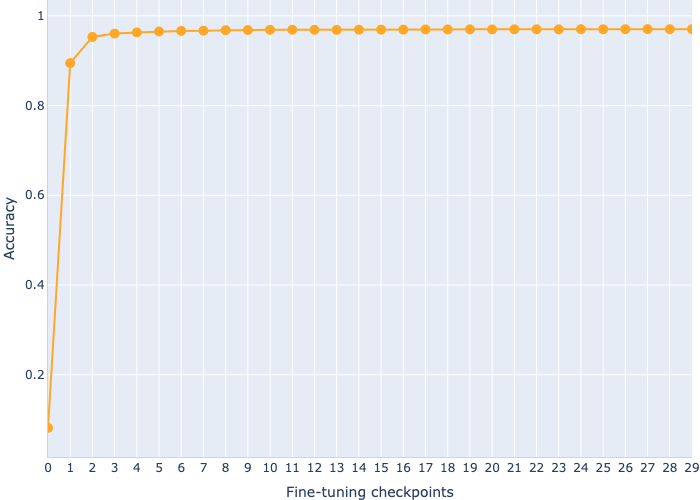}
        \caption{Fine-tuning. Accuracy}
        \label{fig:PoS_Tagging_pos_acc}

    \end{subfigure}

    \vspace{10pt}

    \begin{subfigure}[t]{0.49\textwidth}
        \vskip 0pt
        \includegraphics[width=\textwidth]{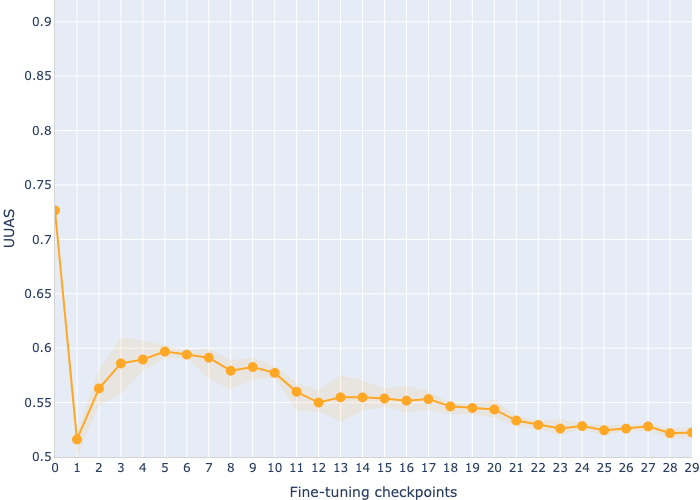}
        \caption{Structural probes tree distance evaluation. \textit{UUAS}}
        \label{fig:PoS_Tagging_distance_uuas}
    \end{subfigure}
    ~
    \begin{subfigure}[t]{0.49\textwidth}
        \vskip 0pt
        \includegraphics[width=\textwidth]{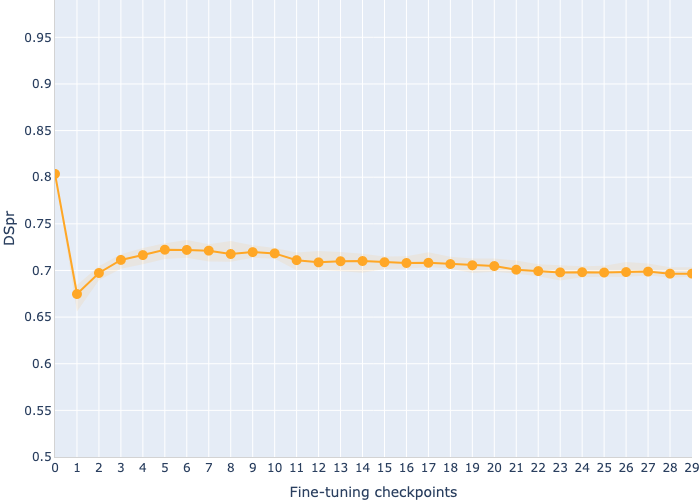}
        \caption{Structural probes tree distance evaluation. \textit{Dspr}}
        \label{fig:PoS_Tagging_distance_spearmanr_mean}
    \end{subfigure}

    \vspace{10pt}

    \begin{subfigure}[t]{0.49\textwidth}
        \vskip 0pt
        \includegraphics[width=\textwidth]{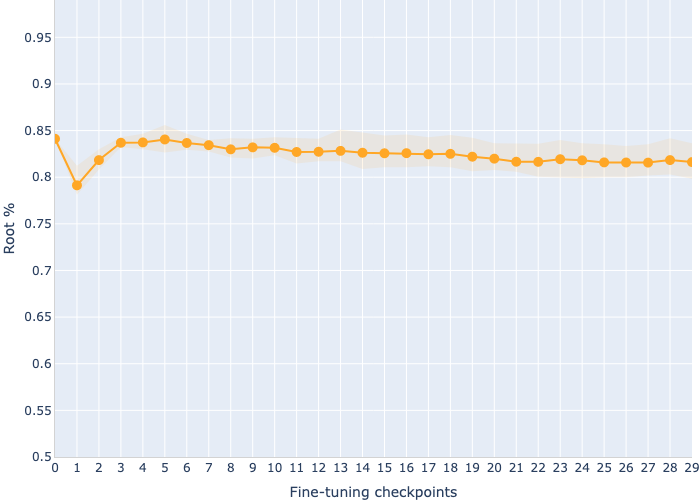}
        \caption{Structural probes tree depth evaluation. \textit{Root \%}}
        \label{fig:PoS_Tagging_depth_root}
    \end{subfigure}
    ~
    \begin{subfigure}[t]{0.49\textwidth}
        \vskip 0pt
        \includegraphics[width=\textwidth]{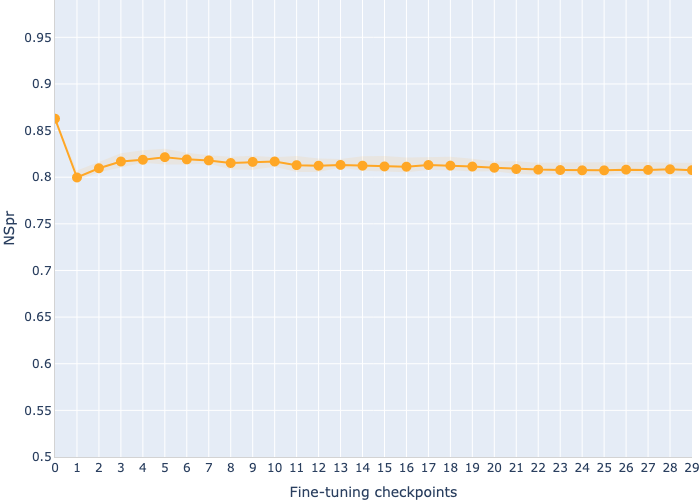}
        \caption{Structural probes tree depth evaluation. \textit{Nspr}.}
        \label{fig:PoS_Tagging_depth_spearmanr_mean}
    \end{subfigure}
    \caption{POS Tagging. Fine-tuning \& probing metrics evolution.}\label{fig:PoS_Tagging}
\end{figure*}

\begin{figure*}
    \centering
    \begin{subfigure}[t]{0.49\textwidth}
        \vskip 0pt
        \paragraph{Dependency parsing with PTB SD} shows a steep learning curve (Figure \ref{fig:Dependency_parsing_PTB_SD_parsing_las}), reaching a performance of 0.90 LAS on the third checkpoint, up to a final 0.94. All four probing metrics show an important improvement in the first fine-tuning step (Figures \ref{fig:Dependency_parsing_PTB_SD_distance_uuas}, \ref{fig:Dependency_parsing_PTB_SD_distance_spearmanr_mean}, \ref{fig:Dependency_parsing_PTB_SD_depth_root} and \ref{fig:Dependency_parsing_PTB_SD_depth_spearmanr_mean}), which is preserved along the rest of the process. As the task heavily relies on syntactic information, this improvement intuitively makes sense. Compared to the result of the other dependency parsing experiments, this one show bigger improvements because the probing is trained on the same dataset.
    \end{subfigure}
    ~
    \begin{subfigure}[t]{0.49\textwidth}
        \vskip 0pt
        \includegraphics[width=\textwidth]{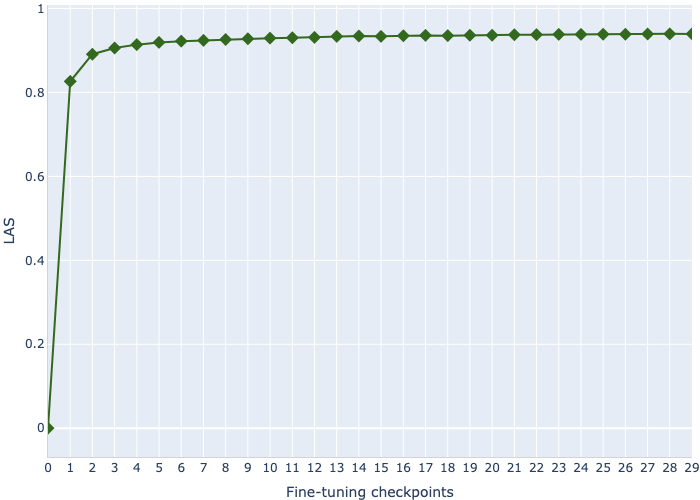}
        \caption{Fine-tuning. LAS}
        \label{fig:Dependency_parsing_PTB_SD_parsing_las}
    \end{subfigure}

    \vspace{10pt}

    \begin{subfigure}[t]{0.49\textwidth}
        \vskip 0pt
        \includegraphics[width=\textwidth]{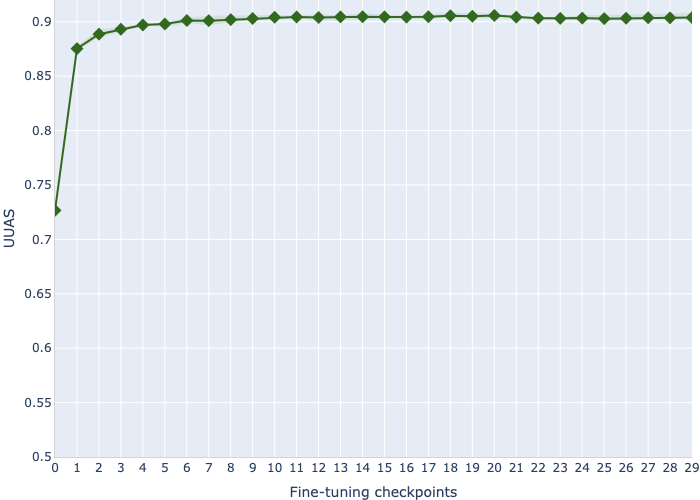}
        \caption{Structural probes tree distance evaluation. \textit{UUAS}}
        \label{fig:Dependency_parsing_PTB_SD_distance_uuas}
    \end{subfigure}
    ~
    \begin{subfigure}[t]{0.49\textwidth}
        \vskip 0pt
        \includegraphics[width=\textwidth]{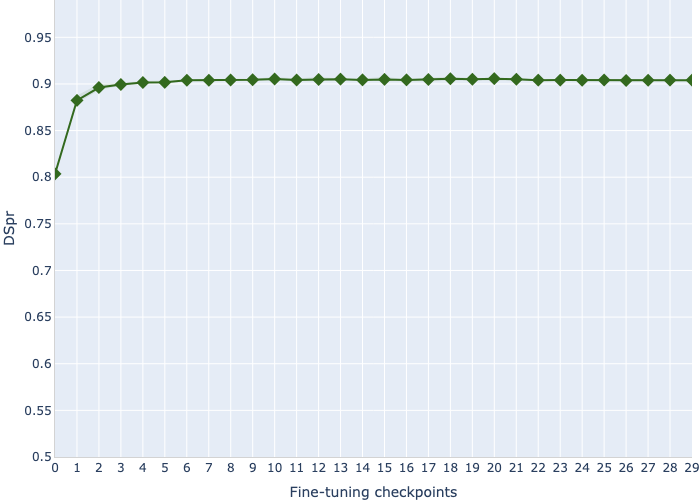}
        \caption{Structural probes tree distance evaluation. \textit{Dspr}.}
        \label{fig:Dependency_parsing_PTB_SD_distance_spearmanr_mean}
    \end{subfigure}

    \vspace{10pt}

    \begin{subfigure}[t]{0.49\textwidth}
        \vskip 0pt
        \includegraphics[width=\textwidth]{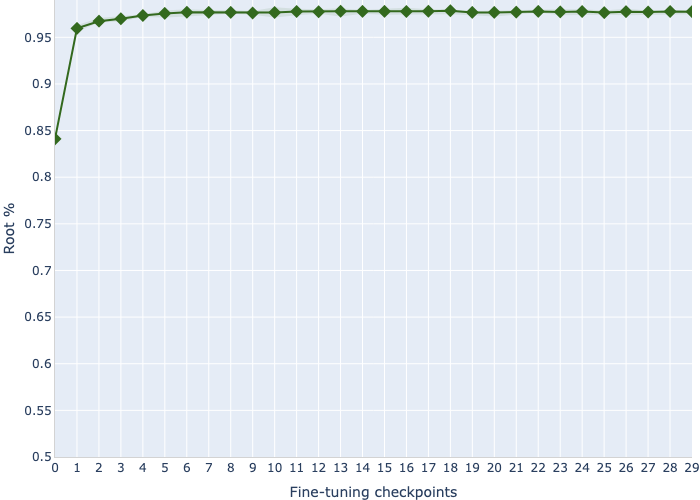}
        \caption{Structural probes tree depth evaluation. \textit{Root \%}}
        \label{fig:Dependency_parsing_PTB_SD_depth_root}
    \end{subfigure}
    ~
    \begin{subfigure}[t]{0.49\textwidth}
        \vskip 0pt
        \includegraphics[width=\textwidth]{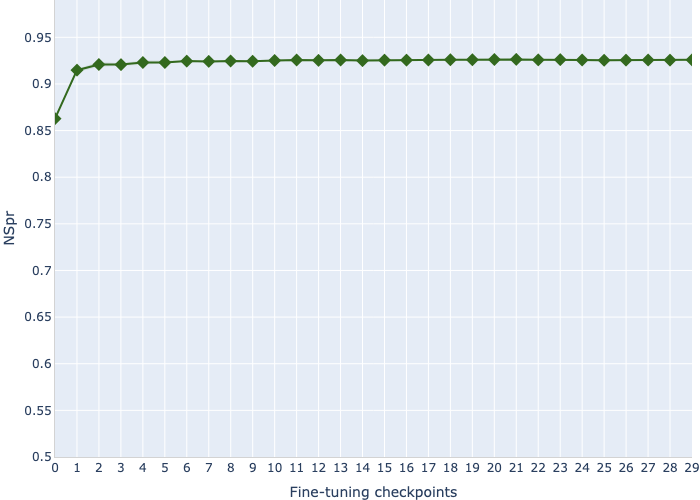}
        \caption{Structural probes tree depth evaluation. \textit{Nspr}}
        \label{fig:Dependency_parsing_PTB_SD_depth_spearmanr_mean}
    \end{subfigure}
    \caption{Dependency Parsing PTB SD. Fine-tuning \& probing metrics evolution.}\label{fig:Dependency_parsing_PTB_SD}
\end{figure*}

\begin{figure*}
    \centering
    \begin{subfigure}[t]{0.49\textwidth}
        \vskip 0pt
        \paragraph{Dependency parsing with EN UD EWT} shows a shallower learning curve than other experiments (Figure \ref{fig:Dependency_parsing_EN_UD_EWT_parsing_las}), as the dataset is significantly smaller than the multilingual and PTB and therefore the models see less examples per checkpoint, ending up with a high performance of 0.9. After an initial drop (probably due to the dataset size, as mentioned before), the probing metrics show a big improvement in the first fine-tuning steps, preserved along the rest of the process (Figures \ref{fig:Dependency_parsing_EN_UD_EWT_distance_uuas}, \ref{fig:Dependency_parsing_EN_UD_EWT_distance_spearmanr_mean}, \ref{fig:Dependency_parsing_EN_UD_EWT_depth_root} and \ref{fig:Dependency_parsing_EN_UD_EWT_depth_spearmanr_mean}). As the task heavily relies on syntactic information, this improvement intuitively makes sense.
    \end{subfigure}
    ~
    \begin{subfigure}[t]{0.49\textwidth}
        \vskip 0pt
        \includegraphics[width=\textwidth]{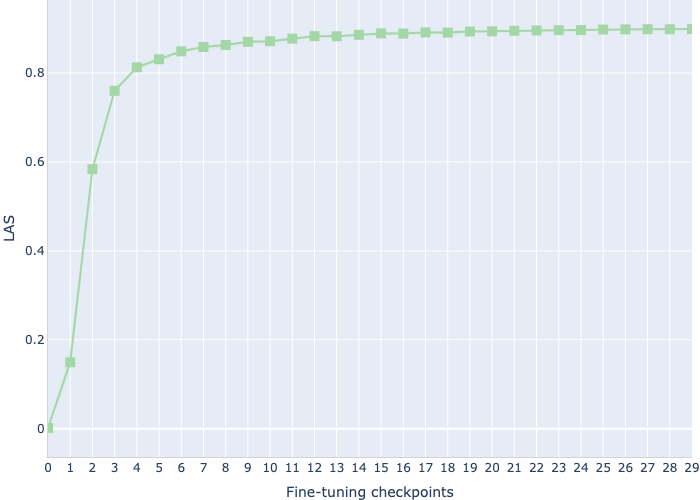}
        \caption{Fine-tuning. LAS}
        \label{fig:Dependency_parsing_EN_UD_EWT_parsing_las}
    \end{subfigure}

    \vspace{10pt}

    \begin{subfigure}[t]{0.49\textwidth}
        \vskip 0pt
        \includegraphics[width=\textwidth]{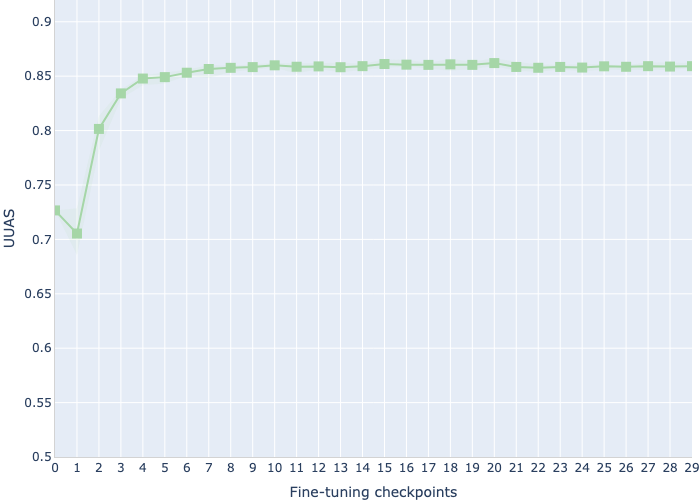}
        \caption{Structural probes tree distance evaluation. \textit{UUAS}}
        \label{fig:Dependency_parsing_EN_UD_EWT_distance_uuas}
    \end{subfigure}
    ~
    \begin{subfigure}[t]{0.49\textwidth}
        \vskip 0pt
        \includegraphics[width=\textwidth]{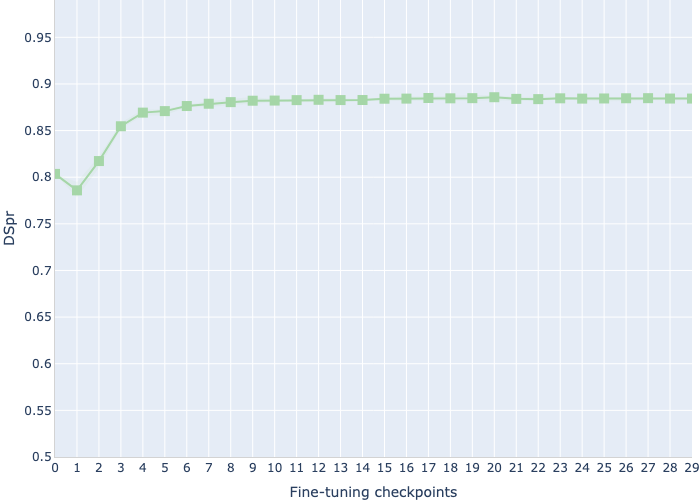}
        \caption{Structural probes tree distance evaluation. \textit{Dspr}.}
        \label{fig:Dependency_parsing_EN_UD_EWT_distance_spearmanr_mean}
    \end{subfigure}

    \vspace{10pt}

    \begin{subfigure}[t]{0.49\textwidth}
        \vskip 0pt
        \includegraphics[width=\textwidth]{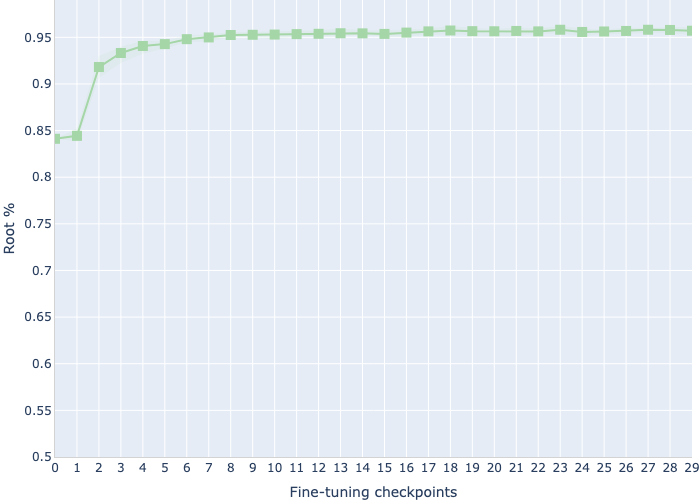}
        \caption{Structural probes tree depth evaluation. \textit{Root \%}}
        \label{fig:Dependency_parsing_EN_UD_EWT_depth_root}
    \end{subfigure}
    ~
    \begin{subfigure}[t]{0.49\textwidth}
        \vskip 0pt
        \includegraphics[width=\textwidth]{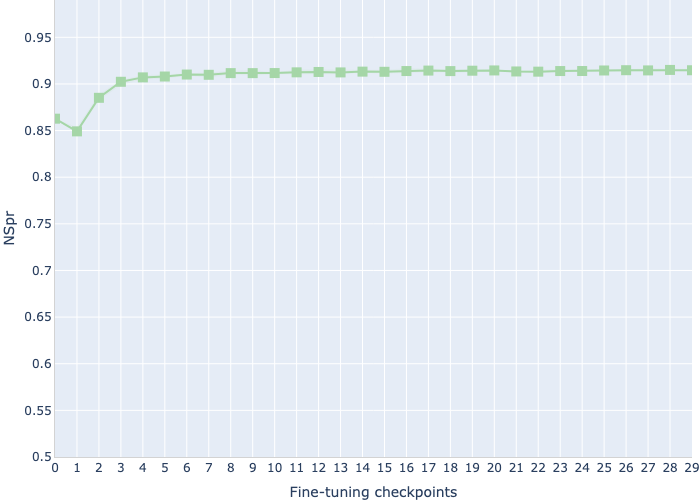}
        \caption{Structural probes tree depth evaluation. \textit{Nspr}}
        \label{fig:Dependency_parsing_EN_UD_EWT_depth_spearmanr_mean}
    \end{subfigure}

    \caption{Dependency Parsing EN UD EWT. Fine-tuning \& probing metrics evolution.}\label{fig:Dependency_parsing_EN_UD_EWT}
\end{figure*}

\begin{figure*}
    \centering
    \begin{subfigure}[t]{0.49\textwidth}
        \vskip 0pt
        \paragraph{Multilingual dependency parsing} shows a steeper learning curve than dependency parsing with EN UD EWT, as it is trained with a larger dataset (Figure \ref{fig:Dependency_parsing_UD_multilingual_parsing_las}), reaching a performance of 0.87 in LAS. All four probing metrics show a big improvement in the first fine-tuning step, preserved along the rest of the process (Figures \ref{fig:Dependency_parsing_UD_multilingual_distance_uuas}, \ref{fig:Dependency_parsing_UD_multilingual_distance_spearmanr_mean}, \ref{fig:Dependency_parsing_UD_multilingual_depth_root} and \ref{fig:Dependency_parsing_UD_multilingual_depth_spearmanr_mean}). As the task heavily relies on syntactic information, this improvement intuitively makes sense.
    \end{subfigure}
    ~
    \begin{subfigure}[t]{0.49\textwidth}
        \vskip 0pt
        \includegraphics[width=\textwidth]{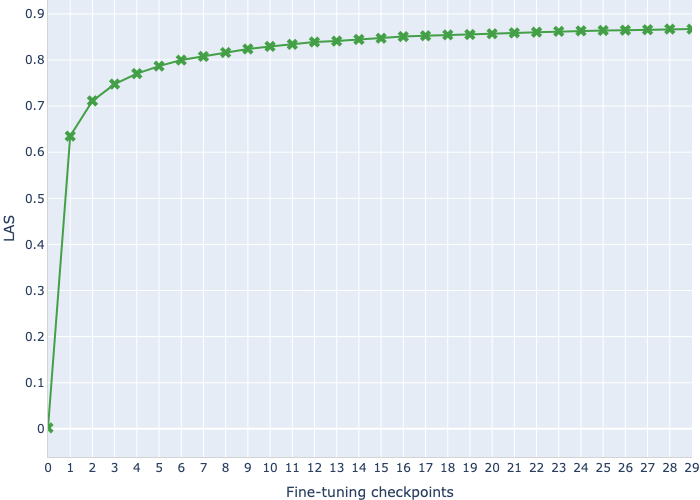}
        \caption{Fine-tuning. LAS}
        \label{fig:Dependency_parsing_UD_multilingual_parsing_las}
    \end{subfigure}

    \vspace{10pt}

    \begin{subfigure}[t]{0.49\textwidth}
        \vskip 0pt
        \includegraphics[width=\textwidth]{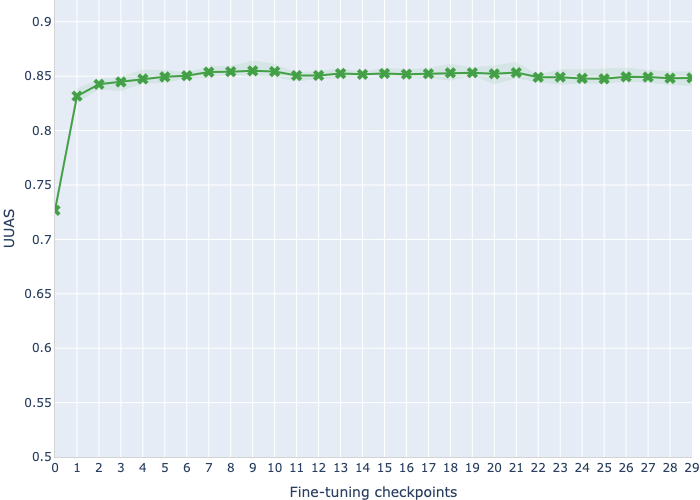}
        \caption{Structural probes tree distance evaluation. \textit{UUAS}}
        \label{fig:Dependency_parsing_UD_multilingual_distance_uuas}
    \end{subfigure}
    ~
    \begin{subfigure}[t]{0.49\textwidth}
        \vskip 0pt
        \includegraphics[width=\textwidth]{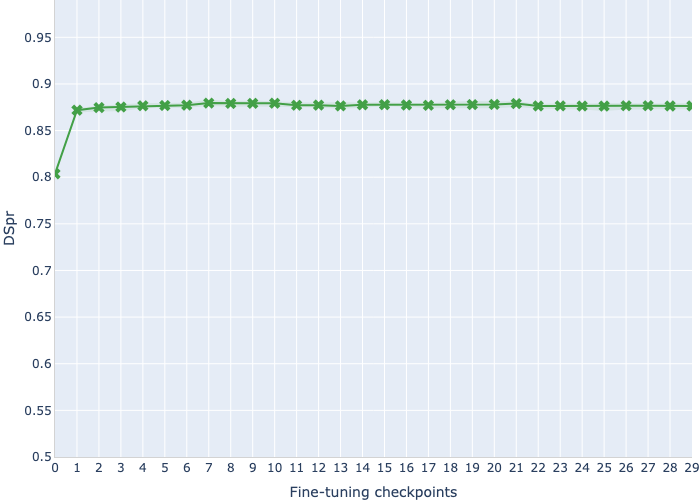}
        \caption{Structural probes tree distance evaluation. \textit{Dspr}.}
        \label{fig:Dependency_parsing_UD_multilingual_distance_spearmanr_mean}
    \end{subfigure}

    \vspace{10pt}

    \begin{subfigure}[t]{0.49\textwidth}
        \vskip 0pt
        \includegraphics[width=\textwidth]{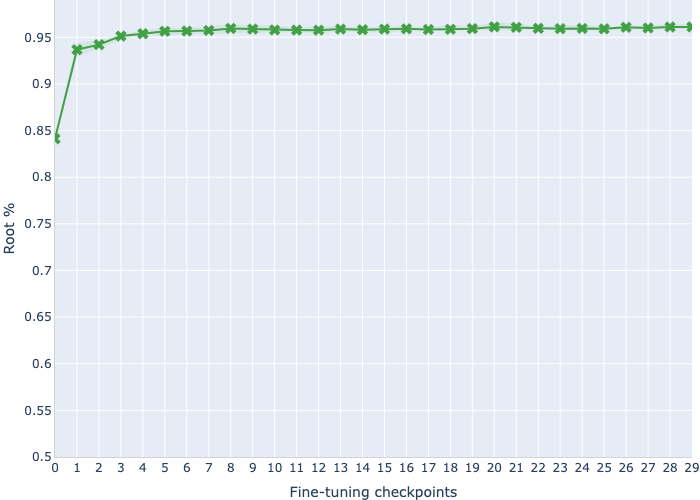}
        \caption{Structural probes tree depth evaluation. \textit{Root \%}}
        \label{fig:Dependency_parsing_UD_multilingual_depth_root}
    \end{subfigure}
    ~
    \begin{subfigure}[t]{0.49\textwidth}
        \vskip 0pt
        \includegraphics[width=\textwidth]{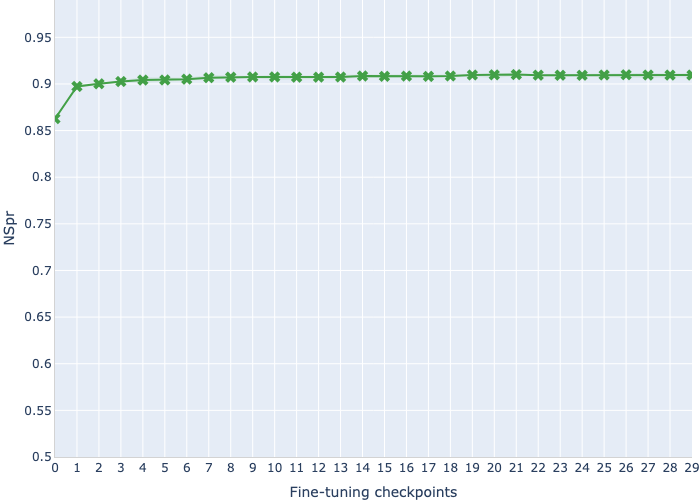}
        \caption{Structural probes tree depth evaluation. \textit{Nspr}}
        \label{fig:Dependency_parsing_UD_multilingual_depth_spearmanr_mean}
    \end{subfigure}

    \caption{Dependency Parsing UD Multilingual. Fine-tuning \& probing metrics evolution.}\label{fig:Dependency_parsing_UD_multilingual}
\end{figure*}

\begin{figure*}
    \centering
    \begin{subfigure}[t]{0.49\textwidth}
        \vskip 0pt
        \paragraph{Constituency parsing} fine-tuning follows a steep curve, quickly reaching an Accuracy of 0.87 that is further improved to 0.9 in the last checkpoint (Figure \ref{fig:Constituency_parsing_pap_acc}). All four probing metrics show a big improvement in the first fine-tuning steps, preserved along the rest of the process (Figures \ref{fig:Constituency_parsing_distance_uuas}, \ref{fig:Constituency_parsing_distance_spearmanr_mean}, \ref{fig:Constituency_parsing_depth_root} and \ref{fig:Constituency_parsing_depth_spearmanr_mean}). As the task heavily relies on syntactic information, this improvement intuitively makes sense. Interestingly, even though the task is modeled as a sequence labeling problem, the probe performs similarly to the dependency parsing tasks, suggesting that the structure of syntax trees emerges in such models even when no tree is explicitly involved in the task.
    \end{subfigure}
    ~
    \begin{subfigure}[t]{0.49\textwidth}
        \vskip 0pt
        \includegraphics[width=\textwidth]{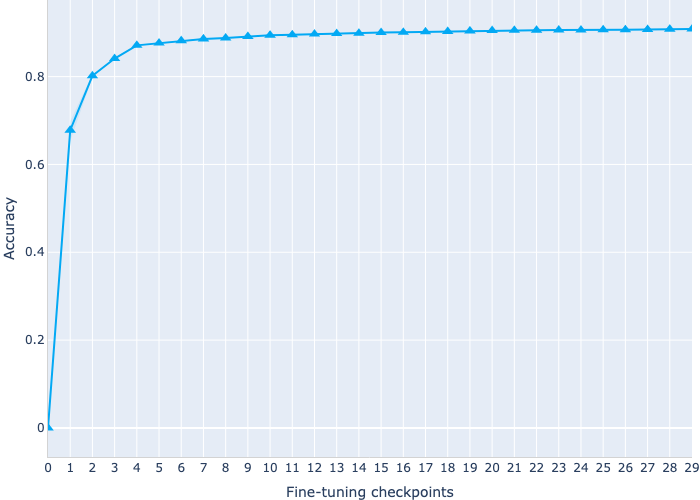}
        \caption{Fine-tuning. Accuracy}
        \label{fig:Constituency_parsing_pap_acc}
    \end{subfigure}

    \vspace{10pt}

    \begin{subfigure}[t]{0.49\textwidth}
        \vskip 0pt
        \includegraphics[width=\textwidth]{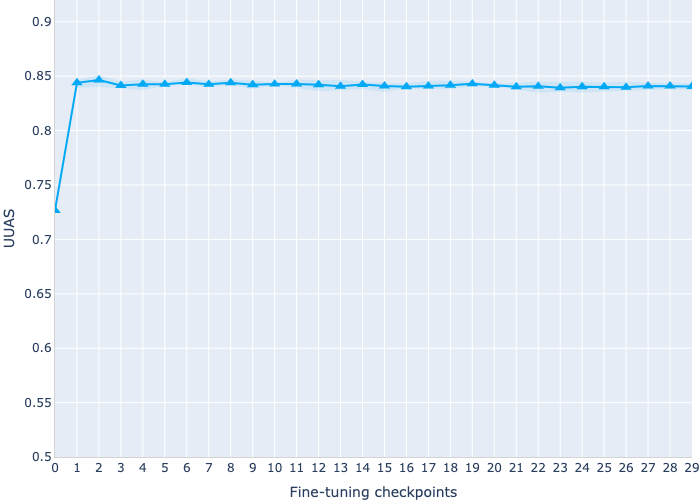}
        \caption{Structural probes tree distance evaluation. \textit{UUAS}}
        \label{fig:Constituency_parsing_distance_uuas}
    \end{subfigure}
    ~
    \begin{subfigure}[t]{0.49\textwidth}
        \vskip 0pt
        \includegraphics[width=\textwidth]{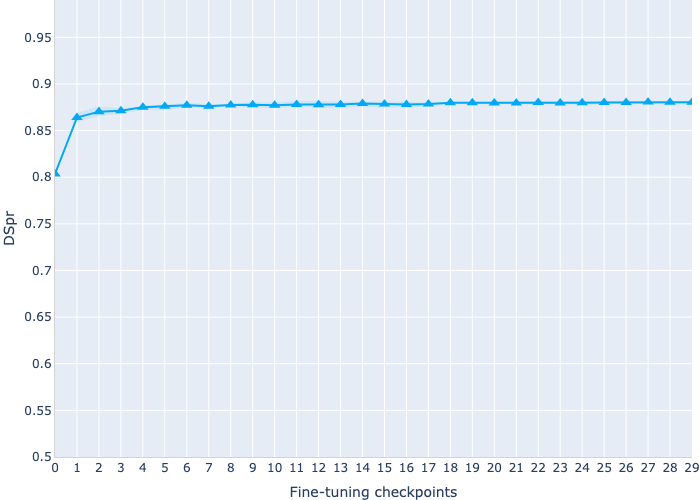}
        \caption{Structural probes tree distance evaluation. \textit{Dspr}.}
        \label{fig:Constituency_parsing_distance_spearmanr_mean}
    \end{subfigure}

    \vspace{10pt}

    \begin{subfigure}[t]{0.49\textwidth}
        \vskip 0pt
        \includegraphics[width=\textwidth]{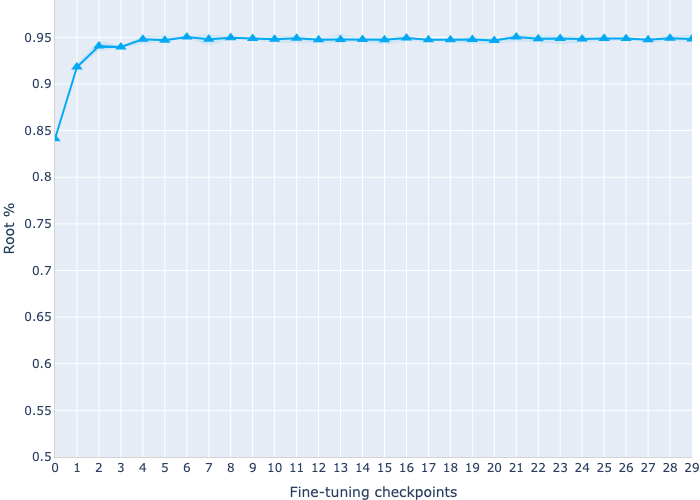}
        \caption{Structural probes tree depth evaluation. \textit{Root \%}}
        \label{fig:Constituency_parsing_depth_root}
    \end{subfigure}
    ~
    \begin{subfigure}[t]{0.49\textwidth}
        \vskip 0pt
        \includegraphics[width=\textwidth]{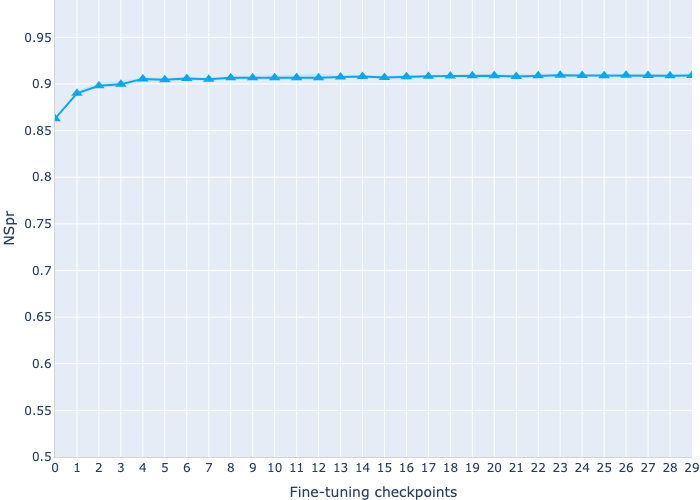}
        \caption{Structural probes tree depth evaluation. \textit{Nspr}}
        \label{fig:Constituency_parsing_depth_spearmanr_mean}
    \end{subfigure}
    \caption{Constituent Parsing. Fine-tuning \& probing metrics evolution.}\label{fig:Constituent_parsing}
\end{figure*}

\begin{figure*}
    \centering
    \begin{subfigure}[t]{0.49\textwidth}
        \vskip 0pt
        \paragraph{Question answering} fine-tuning quickly reaches an F1 score of 0.73 on the first step, which is further improved to 0.88 in the last checkpoint (Figure \ref{fig:SQuAD_squad_f1}). All four probing metrics show a clear loss trend (Figures \ref{fig:SQuAD_distance_uuas}, \ref{fig:SQuAD_distance_spearmanr_mean}, \ref{fig:SQuAD_depth_root} and \ref{fig:SQuAD_depth_spearmanr_mean}). The loss is specially noticeable for \textit{UUAS} and \textit{Root \%}, and more stable for the Spearman correlations, suggesting that even if there is a loss of information it does not impact the distance and depth orderings.
    \end{subfigure}
    ~
    \begin{subfigure}[t]{0.49\textwidth}
        \vskip 0pt
        \includegraphics[width=\textwidth]{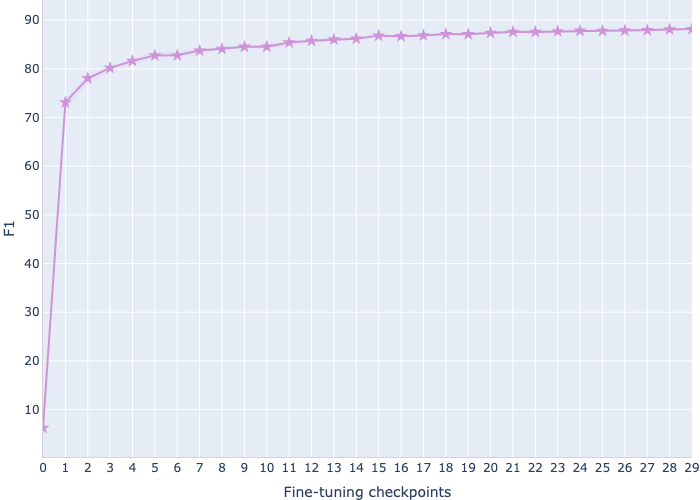}
        \caption{Fine-tuning. F1}
        \label{fig:SQuAD_squad_f1}
    \end{subfigure}

    \vspace{10pt}

    \begin{subfigure}[t]{0.49\textwidth}
        \vskip 0pt
        \includegraphics[width=\textwidth]{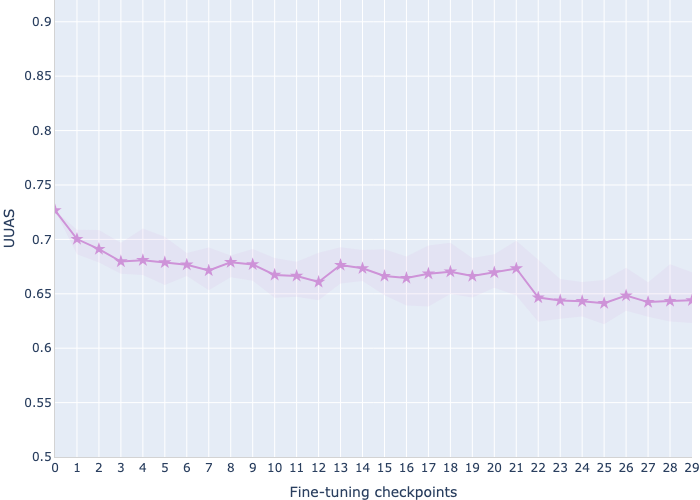}
        \caption{Structural probes tree distance evaluation. \textit{UUAS}}
        \label{fig:SQuAD_distance_uuas}
    \end{subfigure}
    ~
    \begin{subfigure}[t]{0.49\textwidth}
        \vskip 0pt
        \includegraphics[width=\textwidth]{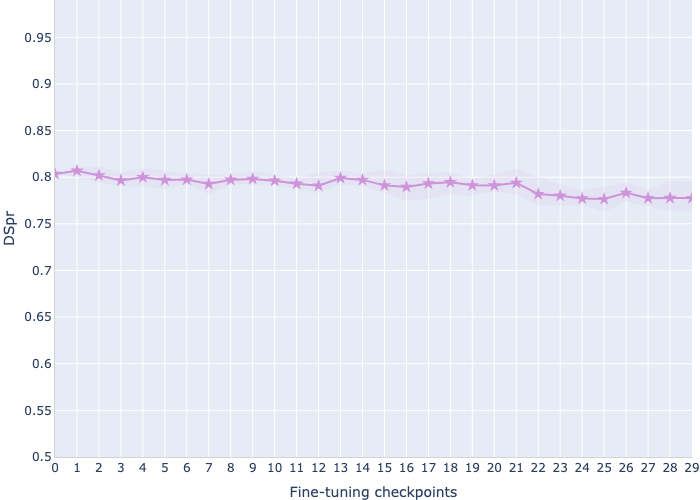}
        \caption{Structural probes tree distance evaluation. \textit{Dspr}.}
        \label{fig:SQuAD_distance_spearmanr_mean}
    \end{subfigure}

    \vspace{10pt}

    \begin{subfigure}[t]{0.49\textwidth}
        \vskip 0pt
        \includegraphics[width=\textwidth]{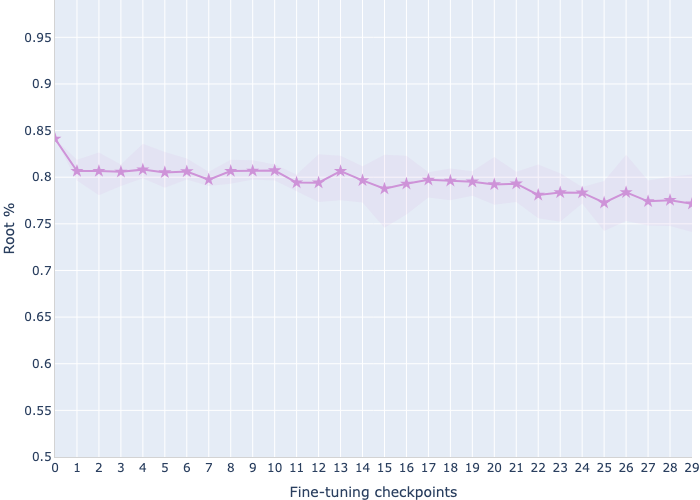}
        \caption{Structural probes tree depth evaluation. \textit{Root \%}}
        \label{fig:SQuAD_depth_root}
    \end{subfigure}
    ~
    \begin{subfigure}[t]{0.49\textwidth}
        \vskip 0pt
        \includegraphics[width=\textwidth]{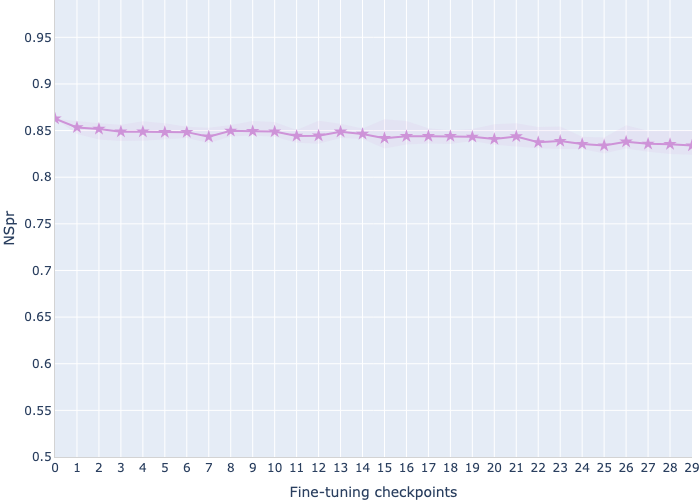}
        \caption{Structural probes tree depth evaluation. \textit{Nspr}}
        \label{fig:SQuAD_depth_spearmanr_mean}
    \end{subfigure}
    \caption{Question Answering. Fine-tuning \& probing metrics evolution.}\label{fig:SQuAD}
\end{figure*}

\begin{figure*}
    \centering
    \begin{subfigure}[t]{1.0\textwidth}
        \vskip 0pt
        \paragraph{Paraphrase identification} fine-tuning starts with an F1 score of 0.81 on the first step that is further improved to 0.90 in the last checkpoint (Figure \ref{fig:MRPC_glue_f1}). Regarding accuracy, after reaching 0.69 on the first checkpoint it follows a shallower curve to a final 0.86 (Figure \ref{fig:MRPC_glue_acc}). All four probing metrics follow a loss trend (Figures \ref{fig:MRPC_distance_uuas}, \ref{fig:MRPC_distance_spearmanr_mean}, \ref{fig:MRPC_depth_root} and \ref{fig:MRPC_depth_spearmanr_mean}). The loss is specially noticeable for \textit{UUAS} and \textit{Root \%}, and more stable for the Spearman correlations, suggesting that even if there is a loss of information it does not impact the distance and depth orderings.
    \end{subfigure}

    \vspace{10pt}

    \begin{subfigure}[t]{0.49\textwidth}
        \vskip 0pt
        \includegraphics[width=\textwidth]{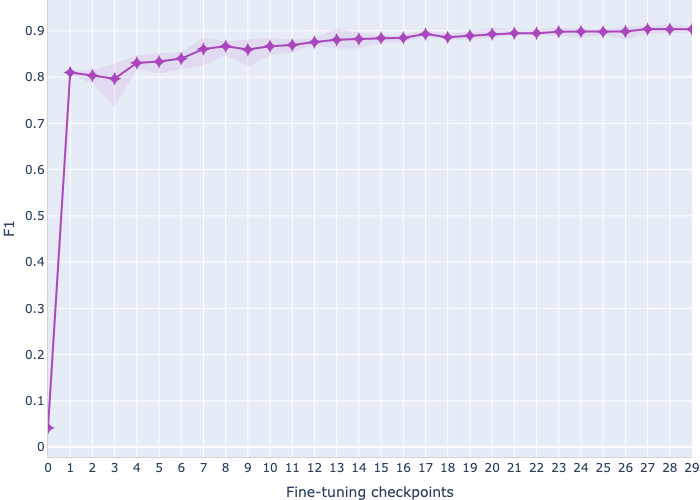}
        \caption{Fine-tuning. F1}
        \label{fig:MRPC_glue_f1}
    \end{subfigure}
    ~
    \begin{subfigure}[t]{0.49\textwidth}
        \vskip 0pt
        \includegraphics[width=\textwidth]{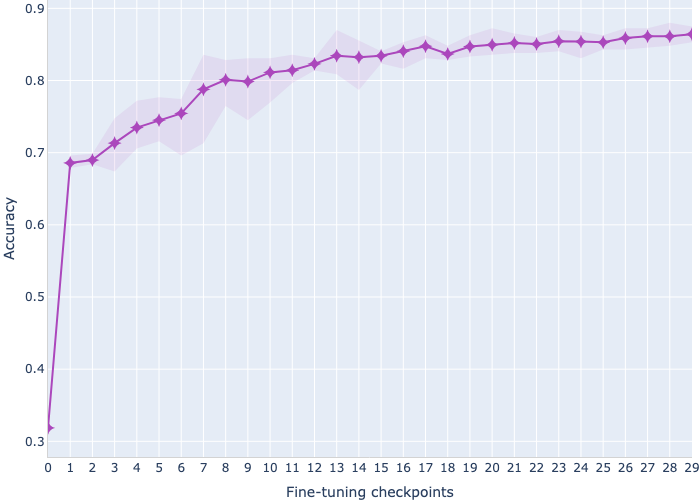}
        \caption{Accuracy}
        \label{fig:MRPC_glue_acc}
    \end{subfigure}

    \vspace{10pt}

    \begin{subfigure}[t]{0.49\textwidth}
        \vskip 0pt
        \includegraphics[width=\textwidth]{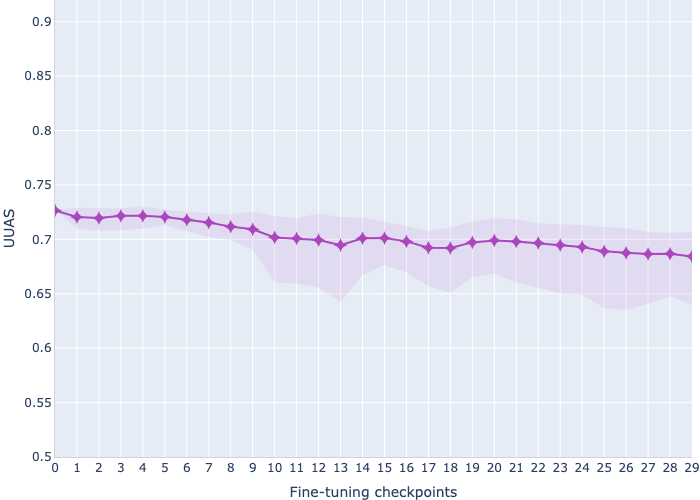}
        \caption{Structural probes tree distance evaluation. \textit{UUAS}}
        \label{fig:MRPC_distance_uuas}
    \end{subfigure}
    ~
    \begin{subfigure}[t]{0.49\textwidth}
        \vskip 0pt
        \includegraphics[width=\textwidth]{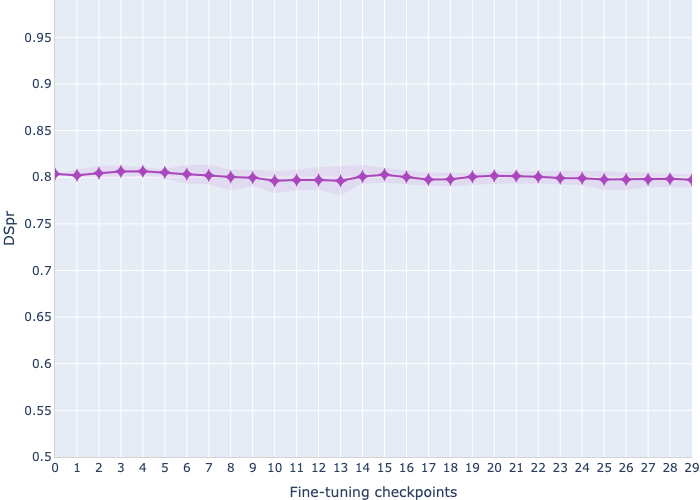}
        \caption{Structural probes tree distance evaluation. \textit{Dspr}.}
        \label{fig:MRPC_distance_spearmanr_mean}
    \end{subfigure}

    \vspace{10pt}

    \begin{subfigure}[t]{0.49\textwidth}
        \vskip 0pt
        \includegraphics[width=\textwidth]{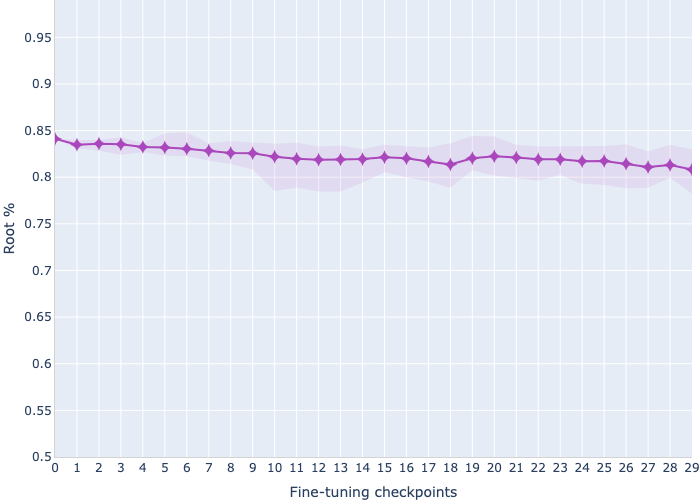}
        \caption{Structural probes tree depth evaluation. \textit{Root \%}}
        \label{fig:MRPC_depth_root}
    \end{subfigure}
    ~
    \begin{subfigure}[t]{0.49\textwidth}
        \vskip 0pt
        \includegraphics[width=\textwidth]{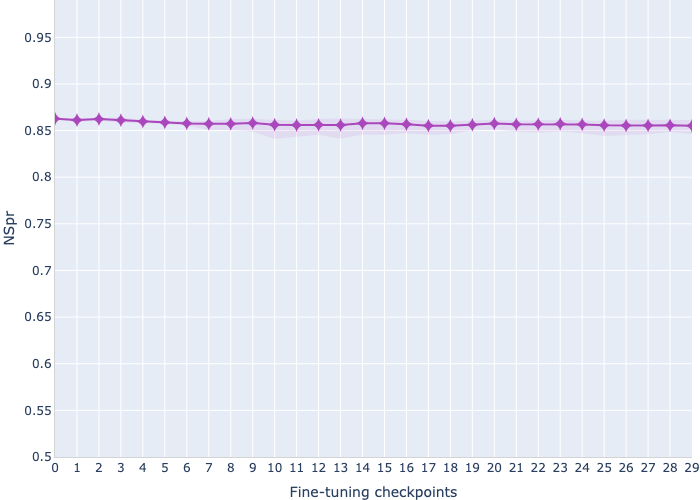}
        \caption{Structural probes tree depth evaluation. \textit{Nspr}}
        \label{fig:MRPC_depth_spearmanr_mean}
    \end{subfigure}
    \caption{Paraphrase identification. Fine-tuning \& probing metrics evolution.}\label{fig:MRPC}
\end{figure*}

\begin{figure*}
    \centering
    \begin{subfigure}[t]{0.49\textwidth}
        \vskip 0pt
        \paragraph{Semantic Role Labeling} fine-tuning follows a steep curve for F1, quickly reaching an F1 score of 0.71 on the first step that is further improved to 0.82 in the last checkpoint (Figure \ref{fig:SRL_f1_measure_overall}). All four probing metrics follow a loss trend (Figures \ref{fig:SRL_distance_uuas}, \ref{fig:SRL_distance_spearmanr_mean}, \ref{fig:SRL_depth_root} and \ref{fig:SRL_depth_spearmanr_mean}). The loss is specially noticeable for \textit{UUAS}, which initially loses around 12 \textit{UUAS} points, and more stable for the Spearman correlations, suggesting that even if there is a loss of information it does not impact the distance and depth orderings.
    \end{subfigure}
    ~
    \begin{subfigure}[t]{0.49\textwidth}
        \vskip 0pt
        \includegraphics[width=\textwidth]{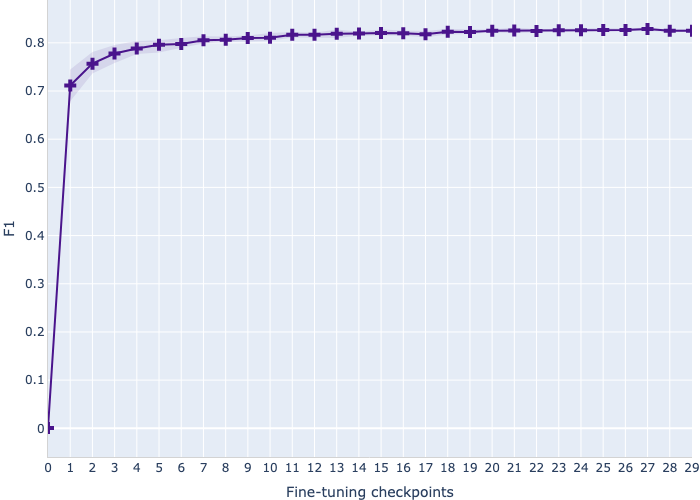}
        \caption{Fine-tuning. F1}
        \label{fig:SRL_f1_measure_overall}
    \end{subfigure}

    \vspace{10pt}

    \begin{subfigure}[t]{0.49\textwidth}
        \vskip 0pt
        \includegraphics[width=\textwidth]{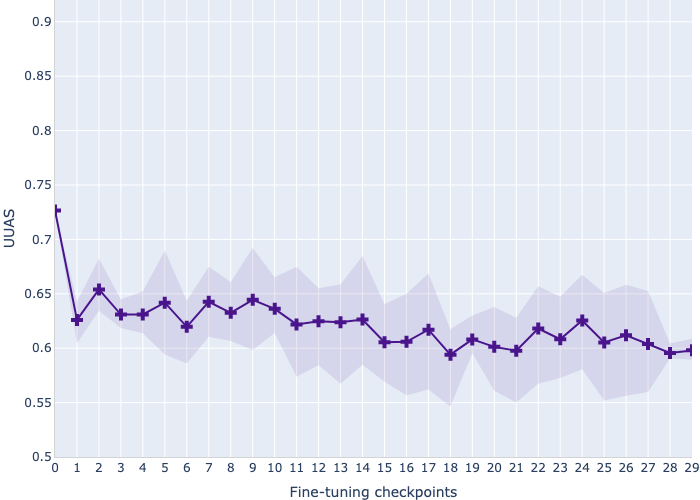}
        \caption{Structural probes tree distance evaluation. \textit{UUAS}}
        \label{fig:SRL_distance_uuas}
    \end{subfigure}
    ~
    \begin{subfigure}[t]{0.49\textwidth}
        \vskip 0pt
        \includegraphics[width=\textwidth]{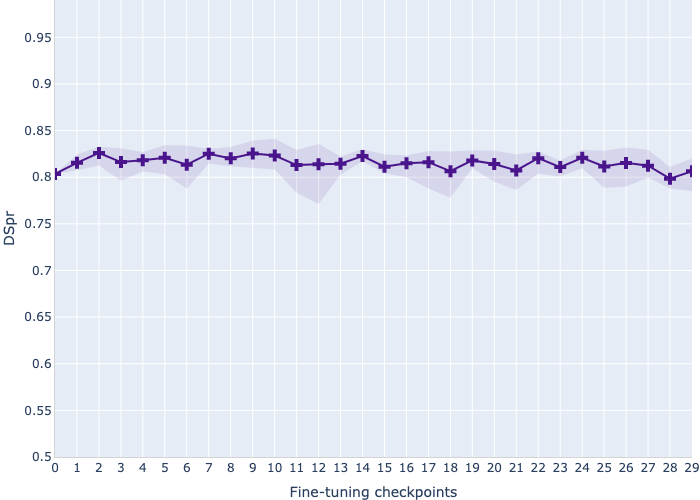}
        \caption{Structural probes tree distance evaluation. \textit{Dspr}.}
        \label{fig:SRL_distance_spearmanr_mean}
    \end{subfigure}

    \vspace{10pt}

    \begin{subfigure}[t]{0.49\textwidth}
        \vskip 0pt
        \includegraphics[width=\textwidth]{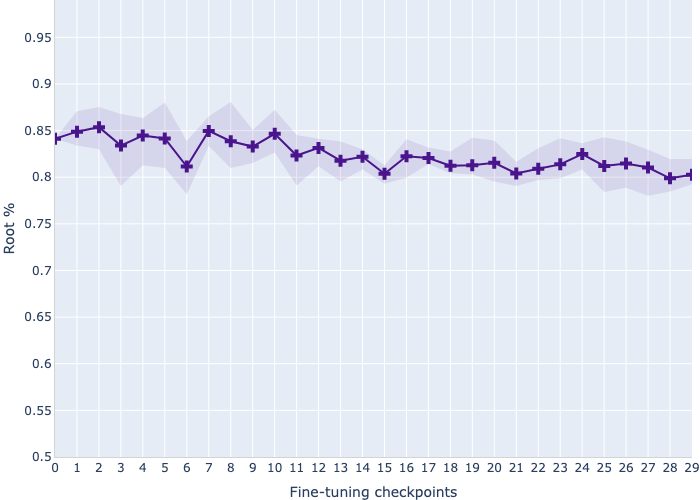}
        \caption{Structural probes tree depth evaluation. \textit{Root \%}}
        \label{fig:SRL_depth_root}
    \end{subfigure}
    ~
    \begin{subfigure}[t]{0.49\textwidth}
        \vskip 0pt
        \includegraphics[width=\textwidth]{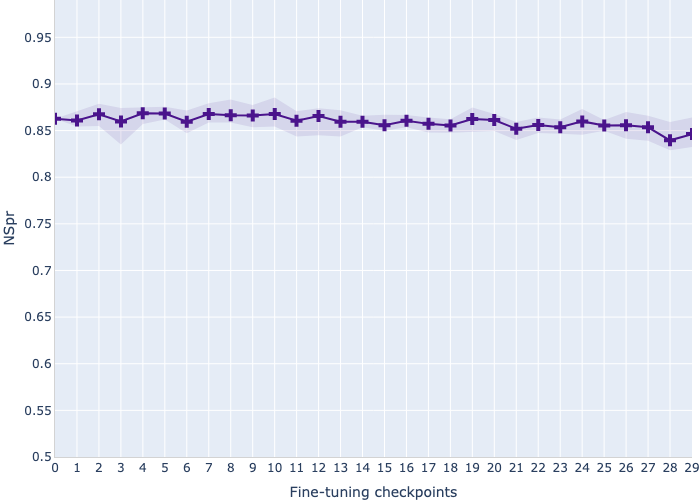}
        \caption{Structural probes tree depth evaluation. \textit{Nspr}}
        \label{fig:SRL_depth_spearmanr_mean}
    \end{subfigure}
    \caption{Semantic Role Labeling. Fine-tuning \& probing metrics evolution.}\label{fig:SRL}
\end{figure*}

Along with the performance curves of the four structural probes metrics (\textit{UUAS}, \textit{Nspr}, \textit{Root \%} and \textit{Dspr}), the following figures include the performance curves of the target tasks and a brief discussion of the results, to help interpretation. Figure \ref{fig:PoS_Tagging} shows the accuracy evolution of PoS tagging. Figures \ref{fig:Dependency_parsing_PTB_SD}, \ref{fig:Dependency_parsing_EN_UD_EWT} and \ref{fig:Dependency_parsing_UD_multilingual} show the Labeled Attachment Score (LAS) of Dependency parsing with PTB SD, EN UD EWT and UD multilingual, respectively. Figure \ref{fig:Constituent_parsing} shows the accuracy evolution of Constituency parsing. Figure \ref{fig:SQuAD} shows the F1 score evolution of Question Answering. Figure \ref{fig:MRPC} shows the F1 score and accuracy evolution of Paraphrase identification. Finally, Figure \ref{fig:SRL} shows the F1 score evolution of Semantic Role Labeling.

\end{document}